\documentclass[sn-mathphys]{sn-jnl}

\jyear{2021}%

\theoremstyle{thmstyleone}%
\theoremstyle{thmstyletwo}%

\theoremstyle{thmstylethree}%

\begin{document}

\title[Article Title]{OCR-RTPS: An OCR-based real-time positioning system for the valet parking}


\author*[1]{\fnm{Zizhang} \sur{Wu}}

\author[1]{\fnm{Xinyuan} \sur{Chen}}
\equalcont{These authors contributed equally to this work.}

\author[1,]{\fnm{Jizheng} \sur{Wang}}
\equalcont{These authors contributed equally to this work.}

\author[1]{\fnm{Xiaoquan} \sur{Wang}}
\equalcont{These authors contributed equally to this work.}

\author[1]{\fnm{Yuanzhu} \sur{Gan}}
\equalcont{These authors contributed equally to this work.}

\author[1]{\fnm{Muqing} \sur{Fang}}
\equalcont{These authors contributed equally to this work.}

\author[2]{\fnm{Tianhao} \sur{Xu}}
\equalcont{These authors contributed equally to this work.}

\affil*[1]{\orgdiv{Zongmu Technology}}
\affil[2]{\orgdiv{Technical University of Braunschweig}}


\abstract{Obtaining the position of ego-vehicle is a crucial prerequisite for automatic control and path planning in the field of autonomous driving. Most existing positioning systems rely on GPS, RTK, or wireless signals, which are arduous to provide effective localization under weak signal conditions. This paper proposes a real-time positioning system based on the detection of the parking numbers as they are unique positioning marks in the parking lot scene. It does not only can help with the positioning with open area, but also run independently under isolation environment. The result tested on both public datasets and self-collected dataset show that the system outperforms others in both performances and applies in practice. In addition, the code and dataset will release later. }




\maketitle

\section{Introduction}\label{sec1}
Classical positioning technologies such as GPS (Global Positioning System) and RTK (Real Time Kinematic) are widespread in autonomous driving \cite{bib1,bib2,bib3,bib4}, but they are limited by the weak signal under isolation scenes \cite{bib5,bib6}, other technologies such as LIDAR, which locates itself by matching the cloud points with the prepared template \cite{bib7,bib8,bib9}, require the high economic and computational cost \cite{bib9}. Therefore, how to obtain the position of vehicles at low expenses in an isolated parking scene, especially the underground parking lot, is an urgent problem to solve.

\begin{figure}[!t]
    \centering
    \includegraphics[width=8cm]{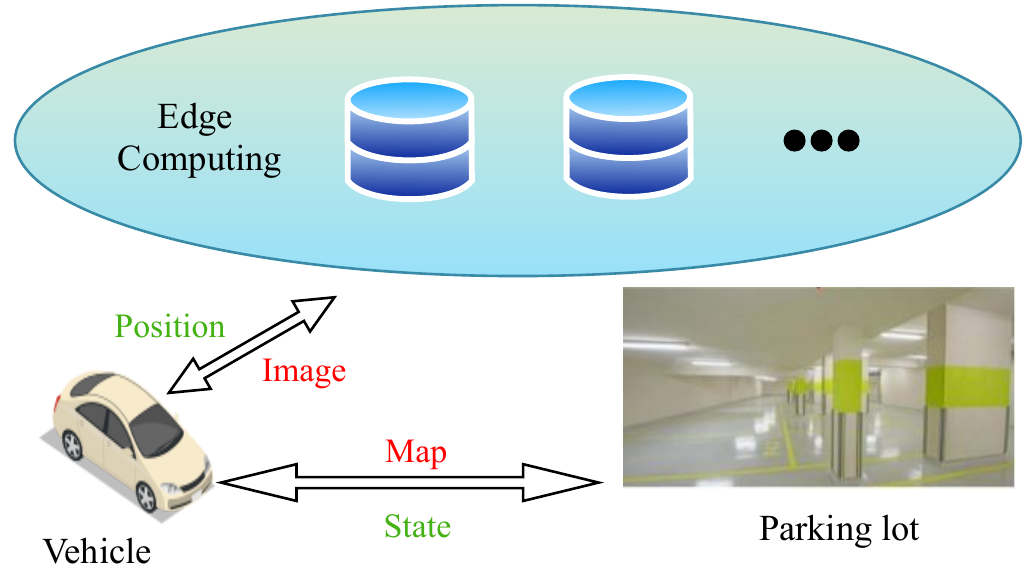}
    \caption{The overview of the proposed method. Firstly, the images captured by the mounted fisheye camera system are applied to edge computing devices to process OCR recognition. Then the edge storage device queries the parking number in an HD(High Definition) semantic map to acquire the coordinate of the parking number. Finally, the parking number position returns to the vehicle.}
    \label{fig:Figure01}
\end{figure}

\begin{figure}[!h]
    \centering
    \includegraphics[width=8.5cm]{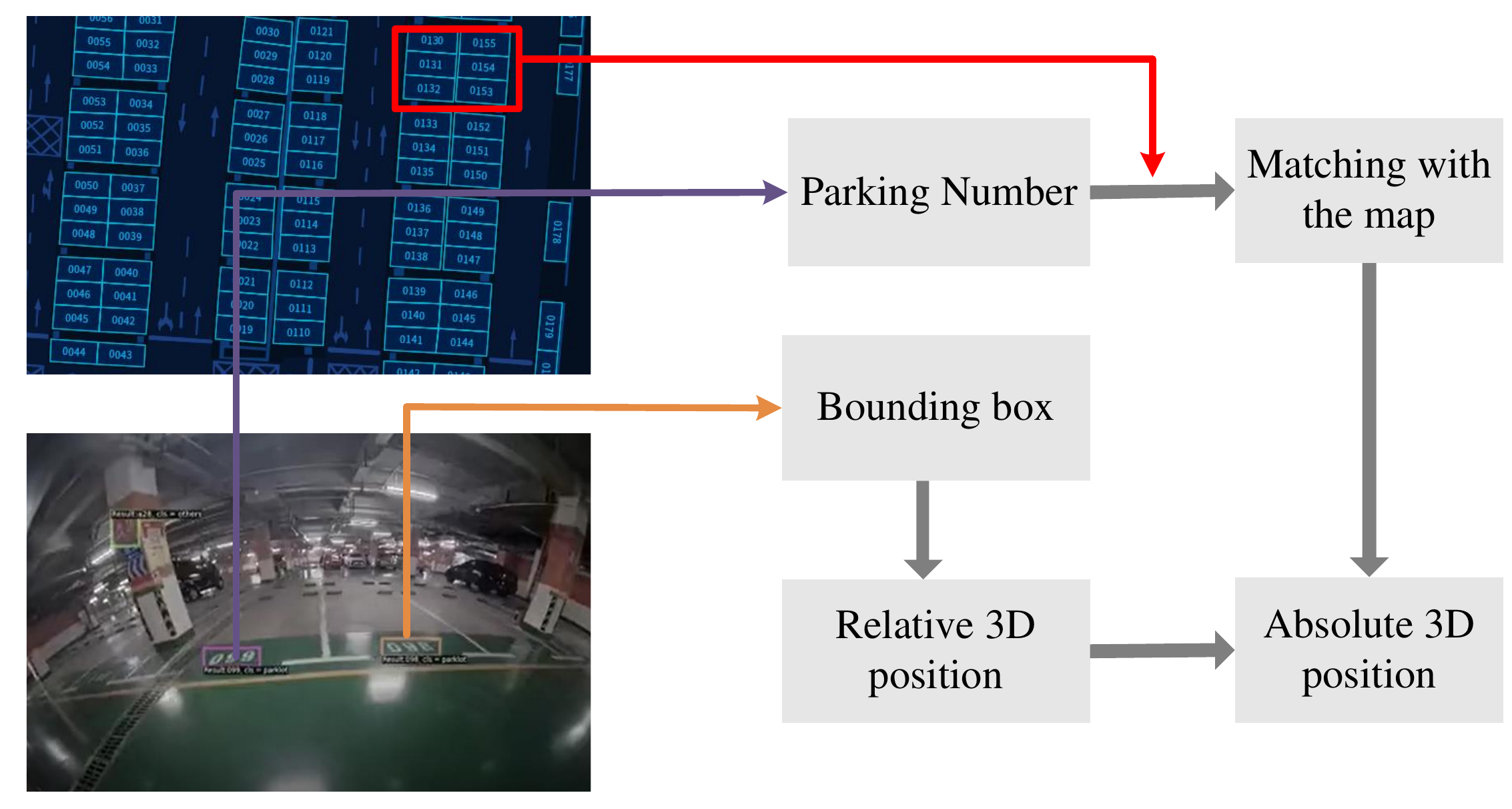}
    \caption{The overall framework of the proposed system. The bounding box of the parking number can be recognized and used to calculate the 3D position relative to the vehicle. The parking number matches with the HD map to acquire its absolute 3D position.}
    \label{fig:Figure02}
\end{figure}

Observing that the parking number is a fixed and unique sign in the parking lot. With the development of Optical Character Recognition (OCR) \cite{bib10,bib11,bib12}, deep learning-based methods \cite{bib13, bib14}  greatly improve the recognition accuracy and the generalization ability to different scenes  \cite{bib15}. These properties make it possible to locate and recognize characters in arbitrary scenes \cite{bib16,bib17}. Considering that, we propose an integrated vehicle positioning system for parking scenes as shown in Fig.\ref{fig:Figure01} and the framework shown in Fig.\ref{fig:Figure02}.

Firstly, we employ a text spotting network (ABCNet \cite{bib13}) to obtain the recognition result of the parking number, the following step is to check the position of the result at the High Definition Map (HD Map) to get its absolute position. Subsequently, The homography matrix is solved using the detected result to obtain the relative position between ego vehicle and parking number. Finally, we combine the relative position with the absolute position of the parking number in the parking lot coordinate system to output the ego vehicle's location. 
Our contributions can be summarized as follows:
\begin{itemize}
\item We first propose an integrated positioning system based on the parking number under the valet parking scene.
\item We establish and annotate a parking number dataset based on the surround-view fisheye camera system to promote the research of related communities.
\item We conducted a large number of comparative experiments and ablation experiments on isolated and open scenarios and verified the superiority of the positioning system.
\end{itemize}
The other parts of the paper are organized as follows. The second section of the paper introduces the related methods of the paper. The third section introduces an overview of the method in this paper. The fourth part describes the new dataset (Parking OCR Dataset) we proposed in detail. The fifth section describes the experimental comparison and analysis of the method proposed in this paper. the sixth section is the discussion of the total positioning system. Finally, the seventh section is about the overall summary of the paper.


\section{Related Work}
\subsection{High Definition Map}
The first High Definition Map(HD Map) was born in 2014 in a Mercedes Benz workshop, for the project Bertha Drive Project \cite{bib26}. The HD Map is usually defined as three layers which are Open Street Map (road network), Lane level map, and Landmarks/road markings map \cite{bib27,bib41,bib42,bib43,bib44,bib45,bib46}. The Open Street Map (road network) \cite{bib44,bib45} is used to plan the overall route, the Lane level map \cite{bib42} helps the vehicle to perceive the current traffic conditions such as roads or obstacles ahead, and the Landmarks/road markings map \cite{bib46} can locate the exact position of the vehicle to change lanes. With the above three layers of functions, we can locate the specific position of the vehicle in the HD map for vehicle perception. Generally, the absolute accuracy of a high precision map is less than one meter, and the relative accuracy reaches the centimeter level usually in 10 to 20 centimeters. In addition, a complete high-precision map should also include road information, such as the number of lanes, traffic lights, crosswalks, lane curvature or slope; rule information, such as speed limits and driving restrictions; real-time information, such as traffic accidents, traffic weather, etc. In a word, high-precision maps play a very important role in the development of autonomous driving.

\subsection{Detection-based methods}
Such methods \cite{bib25,bib28,bib29,bib30,bib31} generally get object detection frames first by popular frameworks under the supervision of word or line-level annotations. CPTN \cite{bib25} and SegLink \cite{bib28} both find text candidates first and then join them in post-processing to get text lines, but require the text to be all horizontal. EAST \cite{bib29} divides text detection into a fully convolutional network stage and an NMS (Non-Maximum Suppression) stage, which can predict rotated and horizontal lines of text. PSENet \cite{bib30} uses ResNet as the backbone to increase the generalization ability to detect text boxes, which can be widely used in various scenarios. DBNet \cite{bib31} proposes differential binarization to generate robust binarized graphs, which greatly simplifies the post-processing process and improves the speed of computation. All of the above methods process text by first detecting and then recognizing, so the result of detection plays a crucial role in the accuracy of recognition.

\subsection{End to end text spotting}
In recent years, deep learning has become one of the most popular directions in artificial intelligence because of its powerful learning and generalization capabilities. Simple scenes based on traditional techniques of the connected domain and sliding window methods can have good results. However, when the image text is distorted, deflected, and the font changes, the detection task of traditional methods completely lags behind that of deep learning methods. End-to-end-based text detection methods should have better robustness and be able to extract high-level features of the scene. With more diverse sample training, deep learning-based OCR techniques have made a breakthrough. In the end-to-end-based methods of detection and recognition, \cite{bib32,bib33,bib34} are highly correlated and complementary, so they can share feature maps in the same model. Li H \cite{bib32} proposes a new method of region feature extraction. This feature extraction method can be well-compatible with text bounding boxes with different original aspect ratios to balance the image distortion. FOTS \cite{bib33} proposes the use of the RoI Rotate module, which bridges the detection and recognition modules. CharNet \cite{bib34} has two branches, the first branch is used for character detection and overall recognition; another branch detects each text within a character(single text or character). This complementary structure avoids recognition errors due to detection errors.

\subsection{Wireless based positioning}
In some indoor scenarios, such as parking lots, and underpasses, it can lead to inaccurate or delayed GPS and RTK positioning. Therefore, in the 21st century where WiFi is widely used, it is very convenient to use wireless signals to locate a car \cite{bib35,bib36,bib37,bib38,bib39,bib40,bib47,bib48,bib49}. WiFi positioning is mainly based on signal strength, fingerprint, angle of arrival, and time of flight. Typically, to determine the target location the distance between the target device and several wifi signal devices is first obtained. Then the relative position of the target device is determined using trilateral measurements, or the angle between the transmitted signals is used to calculate the position of the target. The wifi-based positioning system \cite{bib50} is divided into two types: active positioning \cite{bib51,bib52,bib53,bib54} and passive positioning \cite{bib55,bib56,bib57}. Li W \cite{bib52} proposes a method by utilizing both TOA and RSS measurements. Their purpose is to analyze the difference between  TOA  and  RSS geometric areas to obtain an optimized range of data. Then, these data are introduced into the triangular centroid algorithm to achieve positioning. There are also other methods proposed to improve the accuracy of locating a moving object indoors, such as the TOA/IMU indoor positioning system \cite{bib53}. Oguntala et al. \cite{bib56} use the ranging method of passive RFID receiving signal strength to achieve personnel positioning. The particle filter algorithm analyses and calculates RSSI to get the target position in the indoor environment. Wang X \cite{bib57} purpose CSI phase information is used for indoor fingerprinting and the feasibility of utilizing the calibrated phase information of CSI for indoor positioning is proved. Indoor positioning technology based on WiFi has been widely studied due to its advantages of low cost, wide application, and strong applicability. However, since signal fluctuations may increase localization errors, wifi traces can be blended with other data sources to improve accuracy.
{
\subsection{positioning system under isolated scenes}
Most methods are based on GPS, RTK, WiFi, and other media. In contrast, the positioning method in the isolation scenario does not need to use these media to achieve its positioning, which is worthy of in-depth research. Xia C \cite{bib71} proposed a self-positioning framework without relying on GPS or any other wireless signals. It is proven that the uniform normal distribution changes and two-way information interaction mechanism can achieve the accuracy of the centimeter level, which meets the requirements of autonomous driving cars to achieve real-time stability. Ramasamy P \cite{bib72} established an analysis model, which integrates the geographical reference data from GPS and the signal captured by the visual sensor together to achieve positioning. The simulation results of the study show that compared with the existing methods, the proposed research provides accuracy and faster treatment of improvement. The above studies have effectively improved positioning accuracy in isolation scenarios. However, it rarely uses OCR technology to achieve effective positioning. This Paper will conduct in-depth research on customs in the automatic parking process.}

\section{Methodology}

\subsection{Overall framework}
The proposed system includes a text recognition module, a video-stream-based anomaly filtering module, and a positioning projection module.

To obtain the vehicle's location, we first employ the text recognition module to capture the location, content, and category of text instances(Fig.\ref{fig:Figure02}). Then, these results are sent into the anomaly filtering module to ensure the consistency of positioning. Subsequently, we acquire the parking number's absolute position by matching the detected characters and their corresponding position on the prepared HD Map. In the positioning projection module, the final location is generated by merging the relative distance and orientation between the parking number and the driving vehicle with the absolute location of the parking number.

\subsection{Text recognition module}
\begin{figure}[b]
    \centering
    \includegraphics[width=8.5cm]{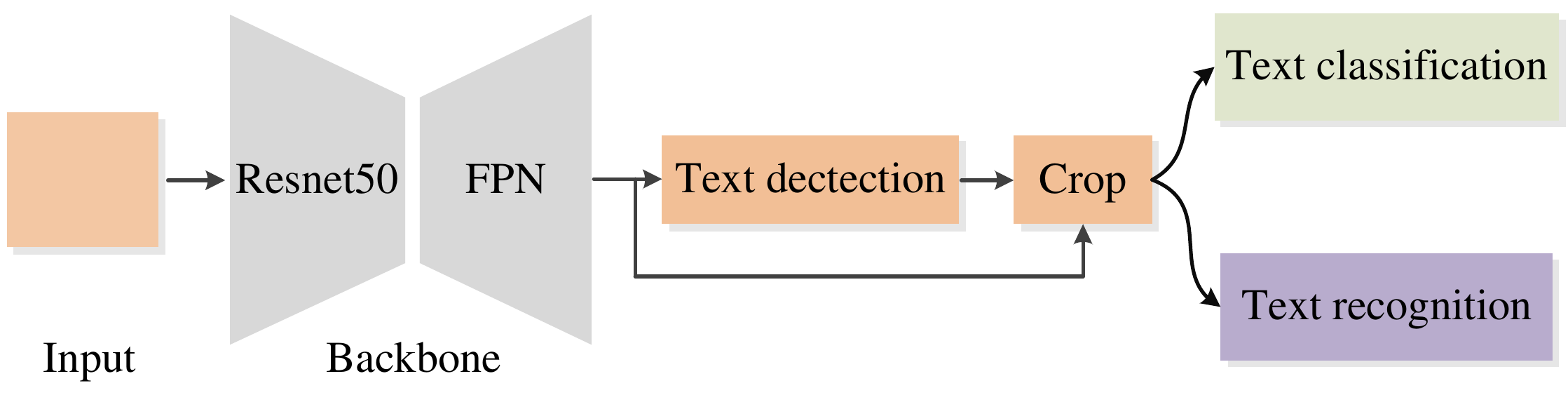}
    \caption{Schematic illustration of our text recognition model.}
    \label{fig:Figure05}
\end{figure}

Fig.\ref{fig:Figure05} illustrates the proposed text recognition module. It is noticeable that any OCR methods can be used in this module, in this paper, We employ ABCNet \cite{bib13} to perform text detection tasks and text recognition tasks, and we stack an additional text classification head to determine if the text instance is a parking number. The input image goes through the backbone (Resnet50 \cite{bib18} \& Feature Pyramid Networks \cite{bib19}) firstly, and the obtained feature maps are fed to the text detection head to acquire the bounding box. Then, to get the category of text instances(parking lot instances, car plate instances, pillar instances, etc.) the predicted bounding boxes are expanded with different ratios to crop the feature maps, this can help the bounding boxes to capture surrounding information and improve the classification performance subsequently. In practice, we take 2 as the expanding ratio. Finally, we feed these expanded feature maps into the classification head and recognition head to get the instance's category and its content. 

\begin{figure}[t]
    \centering
    \includegraphics[width=6cm]{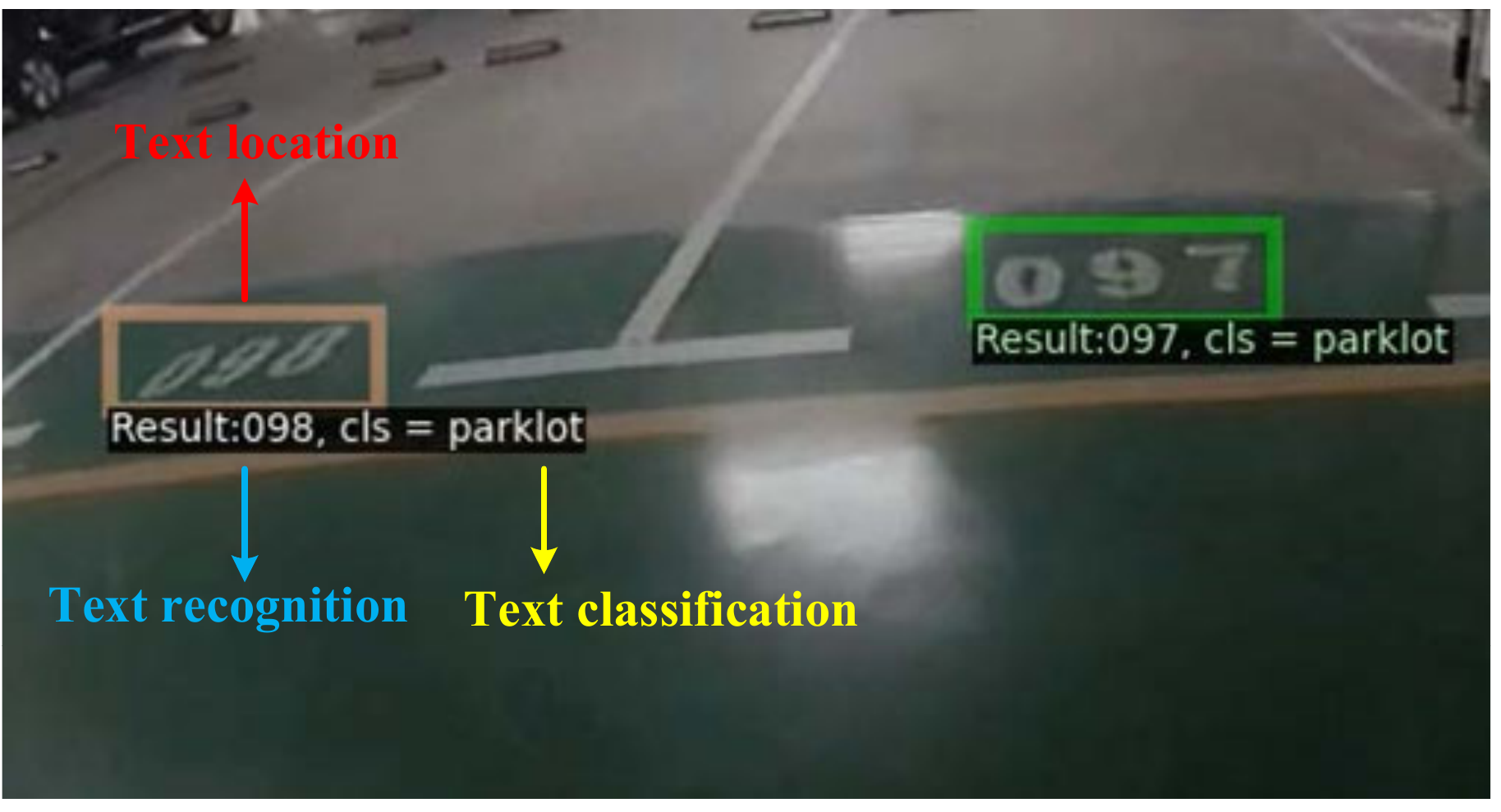}
    \caption{The visualization of the text recognition results. The brown box indicates the location of the leftmost text instance.}
    \label{fig:Figure04}
\end{figure}

Taking the most left text instance shown in Fig.\ref{fig:Figure04} as an example, the text recognition module outputs three results: location, class, and content. The location indicates its coordinates (brown box), the class means the category(parking lot in this example) and the content is the recognition result ("098"). 

\subsection{Anomaly filtering module}
Regarding the failing cases of the text recognition module which are caused by various lighting conditions, occluded parking numbers, different parking angles, etc. We propose an anomaly filtering module to ensure the accuracy and smoothness of localization.

\subsubsection{The robustness of model}
To minimize the anomalies, we first train the OCR model under various conditions (weak/bright lighting, rotated/blocked text). When training data is ideal, it is easy to identify errors in some special circumstances, such as blocking, brightness, blur, etc. The use of data enhancement methods for training data can enhance model robustness, and also improves the generalization ability of the model. Therefore, we apply some techniques such as data enhancement (flipping, resizing, style conversion, etc.). In Table \ref{tabel_volume}, we conduct experimental comparisons with different amounts of data after data augmentation on the Parking OCR dataset.

\subsubsection{The Video-stream based anomaly filter}

\begin{algorithm}
\caption{The anomaly filtering module}\label{algo}
\begin{algorithmic}[1]
\Require video stream
\Ensure video stream's prediction results 
\State Initialize a first-in-first-out queue with the maximum length of 30.
\While{video stream is not over}
    \State 1. Get the prediction results of the current frame.
    \While{Results are not empty}
        \State 1. pull the next result from the queue.
        \State 2. Extract the numbers of the detection result $N$.
        \State 3. Input $N$ to the anomaly filtering algorithm and check if it is an abnormal detection.
            \If{anomaly filtering algorithm returns False }
                \State push the result into the queue
            \Else[anomaly filtering algorithm returns True ]
                \State drop this result
            \EndIf
    \EndWhile
\EndWhile
\end{algorithmic}
\end{algorithm}

This module is proposed to filter out the wrong recognition results produced by the OCR module, it utilizes the prior knowledge that the parking numbers change linearly within a fixed range in the parking lot and the abnormal detection will exceed this range. We first initialize a queue and keep inputting incoming numbers until it is full, since the OCR module has high accuracy, it can promise that most results in the queue are correct. Then, new coming results will be checked if there are in the range using the box plot algorithm\cite{bib20}.
As shown in Algorithm \ref{algo}, the queue stores the prediction results of the previous 30 frames, which ensures the continuity of the parking number in the queue. Using the box plot algorithm\cite{bib20}, anomalies are detected by calculating the lower-quartile, median, upper-quartile and $IQR\ (\text{upperquartile}-\text{lowerquartile})$ , and checking if the results in the range of $[ \text{median}-1.5 \times IQR,$ $ median+1.5 \times IQR ]$. Although operations such as swerve will destroy the continuity, it will re-validate within 3 seconds once other parking numbers are recognized.

The advantage of this module lies in the robustness to the anomaly. The anomaly can be effectively filtered out when the most recognized parking numbers are correct (from upper quartile to lower quartile). Besides, the algorithm complexity is $O(n)$, which guarantees real-time requirements.

\subsubsection{The exact matching mechanism}
To filter out those wrongly recognized parking numbers that do not exist in the parking lot, we employ the exact matching mechanism. it requires the recognized parking number to exactly match the parking number on the HD map, and will be abandoned otherwise.

\subsection{Positioning projection module}
In this module, we obtain the physical coordinates of the parking number relative to the ego-vehicle by using the homography matrix \cite{bib21,bib22,bib23}. The homography matrix projects the position from the image to the physical ground plane.

Firstly, to get the homography matrix from the image to the physical ground plane, we assume that the ground is a plane and use homogeneous coordinates to express the process of mapping a point $Q$ on the ground to a point $q$ in the image plane. $Q$ is a three-dimensional coordinate and can be written as ${Q}={\left[ X \ Y \ Z \ 1 \right]}^T$. By removing the $Z$ axis, we can get the following equation:
\begin{equation}
    q = sHQ
    \label{orginal_equ}
\end{equation}
where $q$ is the two-dimensional coordinate: ${q}={\left[ x \ y \ 1 \right]}^T$, $s$ is the scale factor which represents the scale ratio between these two planes. So the homography matrix $H$ is a $3\times3$ matrix.
\begin{equation}
H=sM\left [ r_1 \ r_2 \ t \right ] =\begin{bmatrix}
h_{11}  &h_{12}  &h_{13} \\
h_{21}  &h_{22}  &h_{23} \\
h_{31}  &h_{32}  &h_{33}
\end{bmatrix}
\label{traned_H}
\end{equation}
We can substitute equation (\ref{traned_H}) into equation (\ref{orginal_equ}) to get:
\begin{equation}
\ 
x=\frac{ h_{11}X+h_{12}Y+h_{13} }{ h_{31}X+h_{32}Y+h_{33} } \quad  
y=\frac{ h_{21}X+h_{22}Y+h_{23} }{ h_{31}X+h_{32}Y+h_{33} }
\end{equation}

Finally, the homography matrix $H$ can be solved by at least 4 pairs of points (any 3 points are not collinear), and then its inverse matrix can help to calculate the conversion from image coordinates to physical coordinates.

\begin{figure}[!h]
    \centering
    \includegraphics[width=8cm]{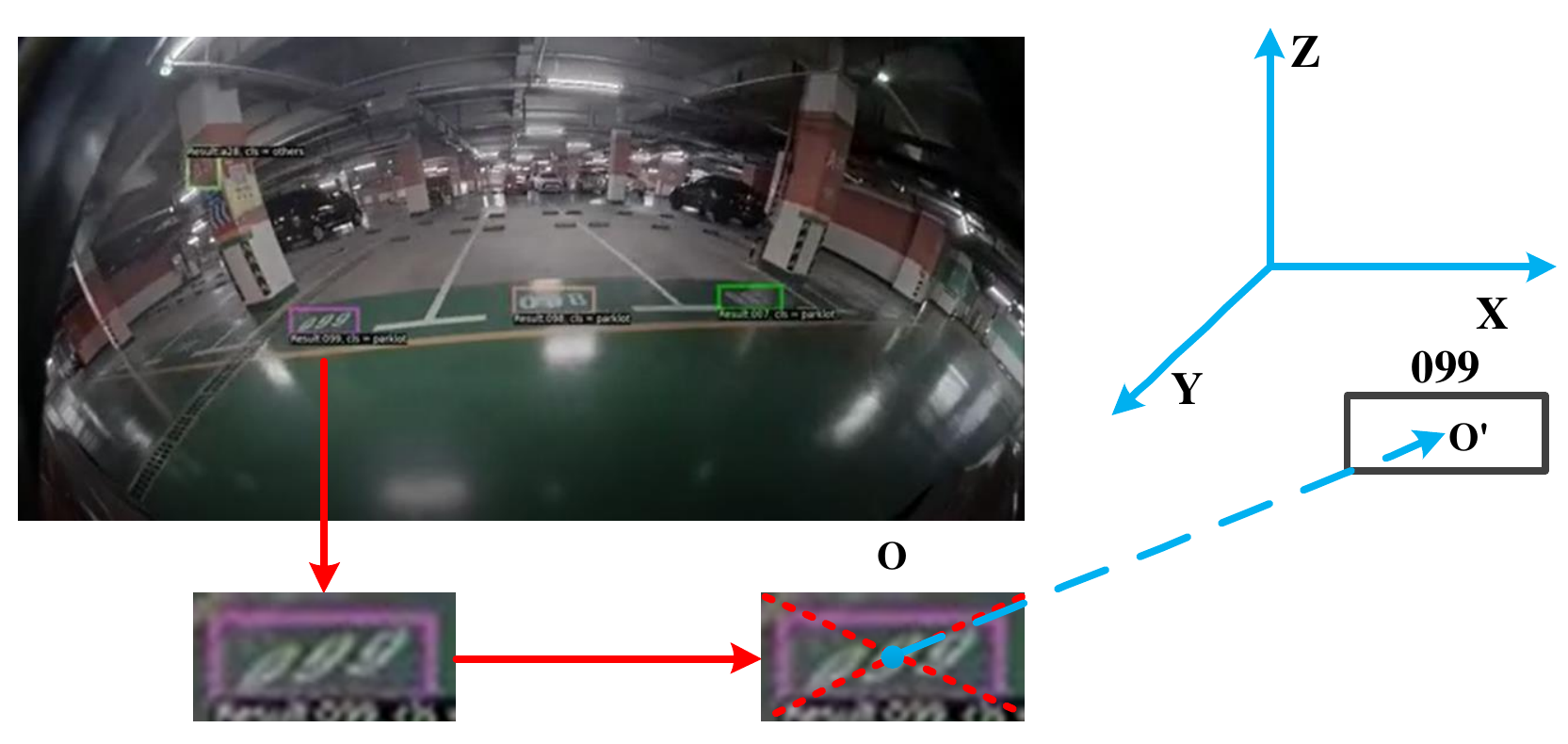}
    \caption{Schematic illustration of parking number projection. The bounding box of the parking number is detected by the text recognition module and its center point on the image is projected into a 3D physical coordinate system by the homography transformation.}
    \label{fig:Figure03}
\end{figure}

By querying the known semantic map, we can obtain the location of the parking number in the physical coordinates. As shown in Fig.\ref{fig:Figure03}, we calculate the center point coordinates $(u,v)$ of text, which is filtered by the anomaly filtering algorithm. After obtaining the relative coordinates between the text center and the vehicle, we can get the absolute position of the ego-vehicle in the physical coordinate system. 

\subsection{Multi camera positioning}
Multi cameras are applied to increase the precision of both absolute positioning and relative positioning. Regarding improving the accuracy of calculating the relative position between parking number and ego vehicle, we use four cameras installed on each side of the vehicle and calculate their relative position to the parking number. The final position is obtained by averaging these results. Besides, this strategy also improves the accuracy of absolute positioning, because the abnormal detection will be removed (As described in section 3.3)  and the more parking numbers we have seen means the more correct parking numbers we get.

{
\section{Parking OCR Dataset}
In this section, we introduce our \textbf{Parking OCR Dataset} in detail, including the data collection, informative statistics, and dataset characteristics.

\subsection{Data Collection}
To ensure the diversity of Parking Text, we collect data from different cities, different parking lots, different periods, and different text cases. 
To be specific, we collect a total of 3 cities, 9 parking lots, two periods (morning and evening), and capture 50 videos. 
These videos last from 1 hour to 6 hours with an average of 2 hours, which includes lots of Parking scene text.
We divide videos into 3 categories: Pillar Text, Park lot Text, and Other Text, according to the videos' content.

We convert the videos into pictures, then pick out the images containing Parking text instances.
For high-quality images, we restrict the visible range of the fisheye camera and further remove distorted and blurred Parking text images.
Additionally, we select the best-quality images, then cover all Parking Text' continuous moving process for redundant annotations.

Finally, we finish collecting the first large-scale fisheye character recognition dataset for parking scenes: Parking OCR Dataset. A total of more than 18 thousand images with the amount of 16744 text instances respectively. Besides, Parking OCR Dataset further performs partition into training, validation, and testing sets with the ratio of 7:2:1.

\subsection{Dataset Description}
Since there is no open-sourced text recognition (OCR) dataset for the parking scene, we establish a large-scale Parking OCR Dataset(Parking Optical Character Recognition Dataset) to show the effectiveness of the proposed system and boost relevant research.  As shown in Table \ref{table2}, the proposed dataset includes text instances under various conditions (weak/strong light and blurry text, etc.) that belong to different parking objectives (parking number, floor indicator on the pillar, license plate characters, and parking sign, etc.). The distribution of the three categories in the Parking OCR Dataset is shown in Fig.\ref{fig:dataset_split}. For each text object, its label includes a bounding box, control points of the bounding box, classification, and content. The Parking OCR dataset contains a total of nine parking lot scenes, as shown in Fig.\ref{fig:Figure_dataset}.

\begin{figure}[!h]
    \centering
    \includegraphics[width=8cm]{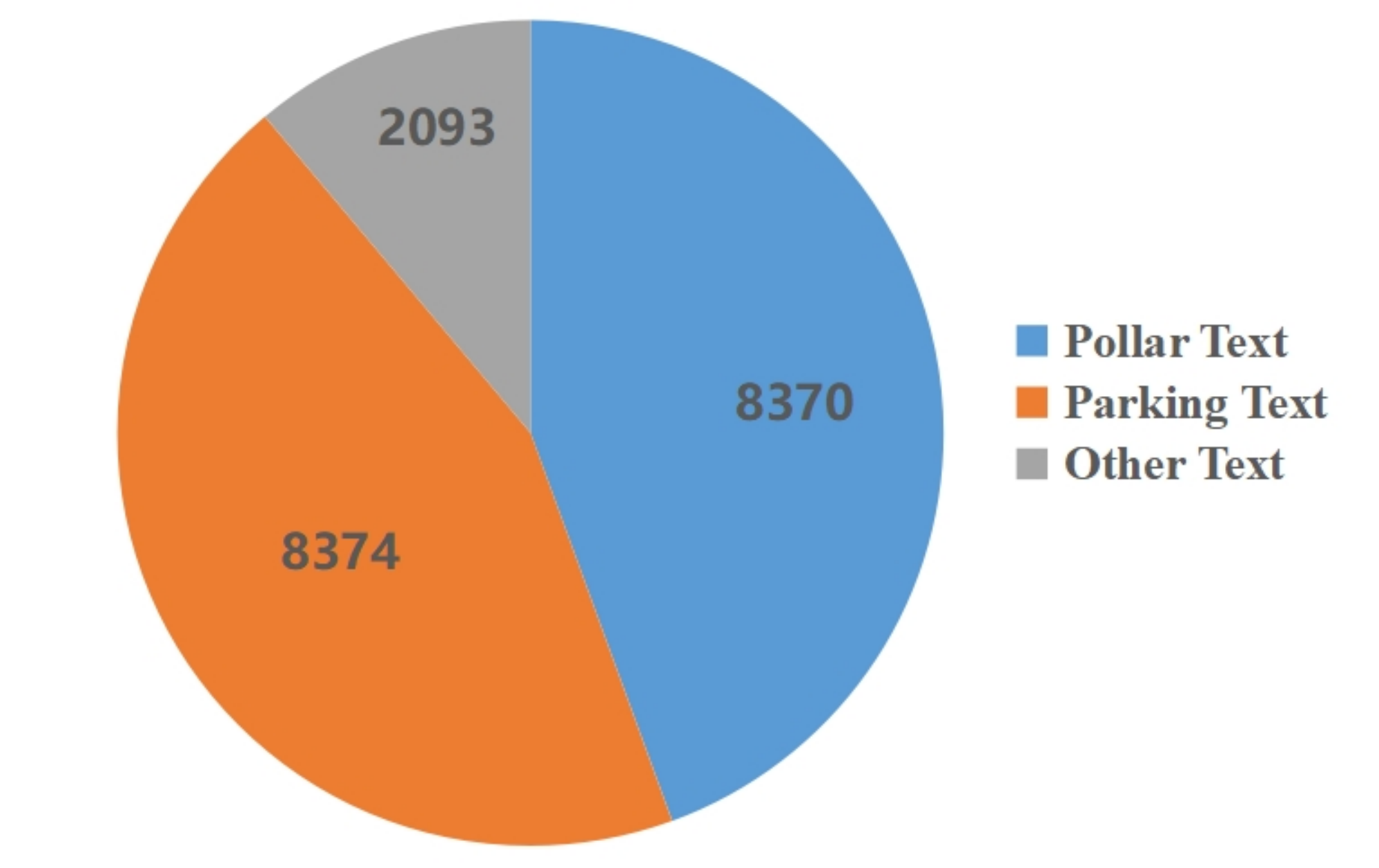}
    \caption{The detailed distribution of Parking OCR Dataset.}
    \label{fig:dataset_split}
\end{figure}

\begin{figure}[!h]
    \centering
    \includegraphics[width=10cm]{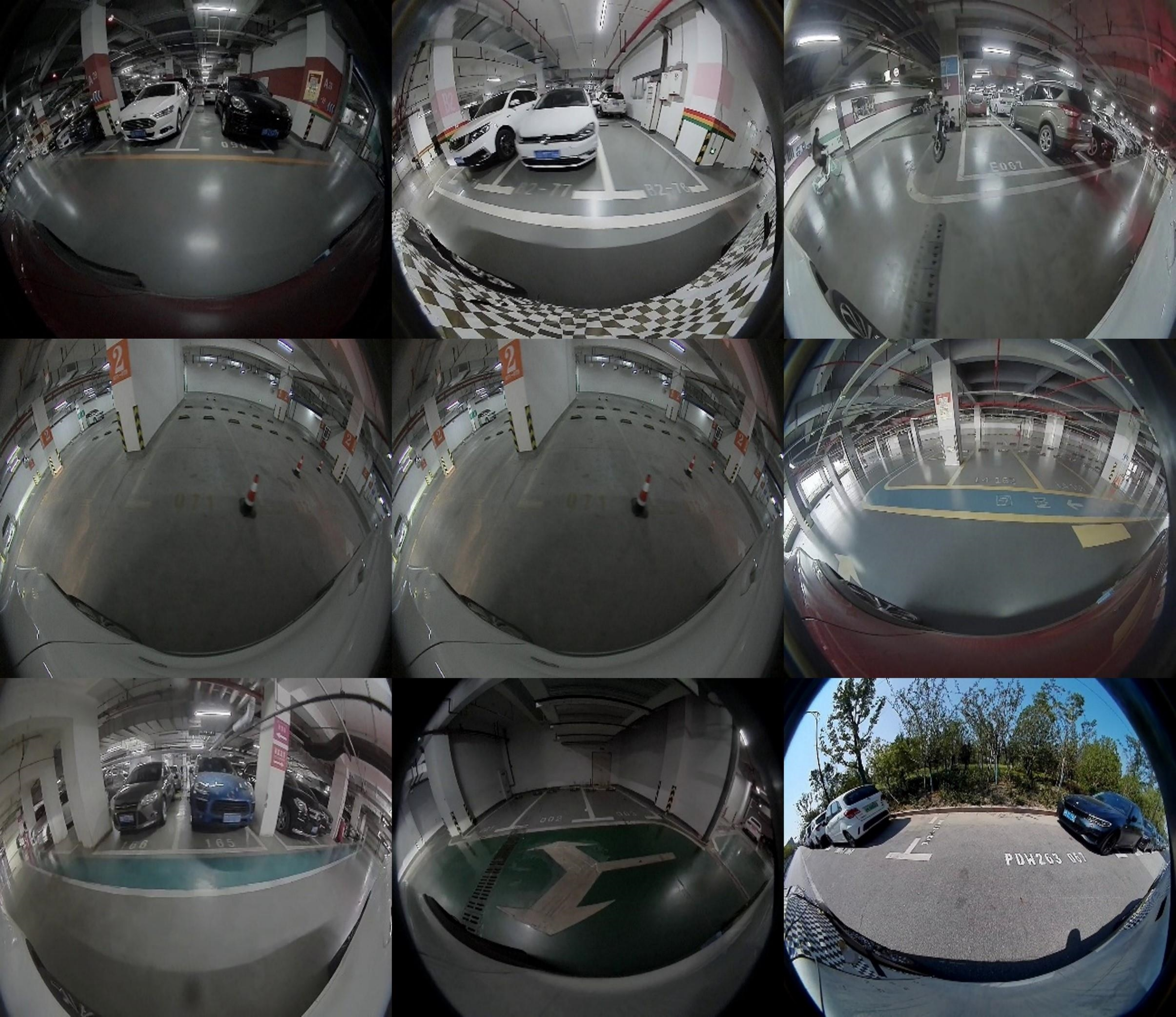}
    \caption{Pictures of nine parking lot scenes in the Parking OCR dataset }
    \label{fig:Figure_dataset}
\end{figure}

\subsection{Dataset Characteristic}
Our Parking OCR Dataset exhibits different from public datasets in image representation style, scenarios, quantity, diversity, and multi-purpose. Below we shall elaborate on the five main characteristics of our \textbf{Parking OCR Dataset}.\\

\noindent{\textbf{Fisheye image representation.}}
Parking OCR Dataset consists of fisheye images, different from the common pinhole images of public datasets. Fisheye images provide a larger field-of-view (FoV), which is more suitable for close-range text detection.\\

\noindent{\textbf{Specifically for parking scenarios.}}
Parking OCR Dataset focuses on text detection in parking scenarios, also distinct from natural scenes of public datasets. Environmental conditions in parking scenarios, such as light and opacity, significantly increase the detection difficulty. Concerning a variety of tough text scenarios, Parking OCR Dataset can promote research in dealing with real-world recognition problems.\\

\noindent{\textbf{Great quantity.}}
Our Parking OCR Dataset obtains more than 18 thousand data from more than 100 hours of parking scene video clips(it will continue to expand in the future). We constantly collect diverse parking text scenarios, eventually reaching the goal of over one hundred thousand data.\\

\noindent{\textbf{High quality and diversity.}}
Our Parking OCR Dataset covers 3 cities, 9 parking lots from different periods, and different text cases to enlarge diversity. Besides, we carefully pick out high-quality images with high resolution, ensuring our dataset's superiority.\\

\noindent{\textbf{Multi-purpose.}}
As a video-based dataset, our Parking OCR Dataset's potential not only lies in character recognition but also other vision tasks, such as pixel-wise semantic segmentation, video object detection, and 3D object detection. Therefore, the Parking OCR Dataset is multi-purpose for diverse tasks. 
}
\section{Experimental}

\begin{table*}[!t]
\centering
\caption{The Status of Parking OCR Dataset. "Number of Pictures" represents the total number of Parking OCR Dataset, and "Number of Text Instances" represents the total text number of Parking OCR Dataset.}
\label{table2}
\resizebox{\linewidth}{!}
{
\begin{tabular}{lllll}
\hline
\multirow{6}{*}{\textbf{Parking\_OCR}} & \textbf{Number of Pictures} & \textbf{Number of Text Instances} & \textbf{Number of Labels} & \textbf{Language} \\ \cline{2-5} 
                              & 18837                 & 16744                  & 16744        & English       \\ \cline{2-5} 
                              & \textbf{Bounding Box Shape} & \textbf{Instances Level} & \textbf{Text\_Classfication\_ID}         & \textbf{Text\_Classfication}        \\ \cline{2-5} 
                              & \multirow{3}{*}{Rectangle \& control points} & \multirow{3}{*}{words} & 0  & Pillar Text \\ \cline{4-5} 
                              &                       &                       & 1         & Parklot Text \\ \cline{4-5} 
                              &                       &                       & 2         & Other Text   \\ \hline
\end{tabular}
}

\end{table*}


\subsection{Evaluation Criteria}
To prove the effectiveness of the system, we evaluate it on our Parking OCR Dataset. We use the accuracy and system recognition time as the evaluation metrics of the whole system. The accuracy rate represents the proportion of the total number of correct predictions by the positioning system, and the system recognition time represents the total time from the detection to the recognition and classification of the system, which is very important for real-time evaluation.


\subsection{Experimental setup}
The experimental procedure is settled up under the Ubuntu 18.04.06 system, together with the python3.8.5 and pytorch1.8. We train the Parking OCR Dataset with 120 epochs. The learning rate is 0.12, momentum is 0.9, the learning decay rate is 0.0001, the optimizer is D2SGD, and Batch\_size is set to 32. It is worth noting that this experiment is taken by using the Parking OCR Dataset (which is going to be open-sourced after publication.) since there is no such parking OCR dataset available currently. we split the Parking OCR Dataset into training and testing subsets that both contain the parking number under various conditions (dark/light illumination, clear/blurry characters, and free/occluded numbers) to test the performance of our system. Besides, Parking OCR Dataset further performs partition into training, validation, and testing sets with the ratio of 7:2:1.

\subsection{Ablation study  }

\subsubsection{Different data augmentation quantity ablation experiments}

In Table \ref{tabel_volume}, we conduct experimental comparisons with different amounts of data after data augmentation on the Parking OCR dataset. As can be seen from the table, with the increase of data enhancement, F-measure has gradually increased. Here we only do the expansion analysis after data enhancement, and the latter experiments do not adopt the data enhancement method by default. Therefore, the means of enhancement of data can effectively enhance the performance of the OCR model, and it also indirectly enhances the effect of abnormal filtration.

\begin{table}[htb]
\caption{experimental comparisons with different amounts of data after data augmentation on the Parking OCR dataset.}
\label{tabel_volume}
\begin{tabular}{c|cccccc}
\hline
The amount of data & +0   & +10000 & +20000 & +30000 & +40000 & +50000 \\ \hline
F-measure          & 74.2 & 74.9   & 75.1   & 75.3   & 75.8   & 75.9   \\ \hline
\end{tabular}
\end{table}

\subsubsection{Comparison of ablation of important components}

To show the effectiveness of our methods, we proposed an ablation study to illustrate the performance of each module. We will stretch and analyze the Anomaly Filtering Module (AFM), Multi Cameras Relative Positioning (MCRP), and Multi Cameras AbSolute Positioning (MCAP).

As shown in table \ref{table10}, with AFM, the results can be improved by 43cm, it greatly improves the performance of the OCR module by filtering out most of the false detection. we also compare the performance with and without multi cameras relative positioning module, the result illustrates that it can finetune the vehicle's location and make it more accurate. it increases the precision from 22 to 19 with the AFM module.

According to the result, the absolute positioning performs better than the MCRP module, it promotes the performance from 19cm to 5cm with the base of AFM and MCRP. The reason behind this may include that the absolute position takes a more important weight than the relative position. 

\begin{table}[htbp]
\caption{In the ablation study, AFM stands for Anomaly filtering module as described in section 3.3. RMCP and AMCP stand for relative multi-camera positioning and absolute multi-camera positioning respectively, which is described in section 3.5. Precision is calculated as the distance between the ground truth and the prediction result.}

\label{table10}     
\centering
\begin{tabular}{c|cccc|c}
\hline
\multicolumn{1}{c}{System} &\multicolumn{1}{c}{OCR} & \multicolumn{1}{c}{AFM} & \multicolumn{1}{c}{MCRP}  & \multicolumn{1}{c}{MCAP} & \multicolumn{1}{c}{Precision(cm)}\\
\hline
\multirow{8}{*}{OCR-RTPS} & {\checkmark} & {-} & {-} & {-}  & {65} \\
& {\checkmark} & {\checkmark}  & {-} & {-} & {22} \\
& {\checkmark} & {\checkmark} & {\checkmark} & {-}  & {19} \\
& {\checkmark} & {\checkmark} & {-} & {\checkmark}  & {17} \\
& {\checkmark} & {-} & {\checkmark} & {-}  & {54} \\
& {\checkmark} & {-} & {-} & {\checkmark}  & {34} \\
& {\checkmark} & {-} & {\checkmark} & {\checkmark}  & {25} \\
& {\checkmark} & {\checkmark} & {\checkmark} & {\checkmark}  & {5} \\
\hline
\end{tabular}
\end{table}

\subsection{Result and evaluation}
To demonstrate the effectiveness of the localization system, we will demonstrate its effectiveness and superiority in three aspects: text recognition module, anomaly filtering module, and comparison with other positioning systems, respectively.
\begin{table}[htbp]
\caption{Comparison between different OCR methods. the result is roughly inferred based on the original paper or the provided code. Datasets:IC13\cite{bib58},IC15\cite{bib59},TT\cite{bib60},MLT\cite{bib61},COCO-Text\cite{bib62}, Parking OCR.}

\label{table06}
\resizebox{\linewidth}{!}{   
\begin{tabular}{l|ccc|ccc}
\hline\noalign{\smallskip}
\multicolumn{1}{c}{Methods} & \multicolumn{1}{c}{Dataset} & \multicolumn{1}{c}{F-measure}  & \multicolumn{1}{c}{FPS}  & \multicolumn{1}{c}{Dataset} & \multicolumn{1}{c}{F-measure}  & \multicolumn{1}{c}{FPS}\\
\noalign{\smallskip}\hline\noalign{\smallskip}
{TextBoxes\cite{bib63}} & {SynText800k,IC13,IC15,TT}  & {48.9} & {1.4} &\multirow{10}{*}{Parking OCR}& {51.3} & {1.6}\\
{MaskTextSpotter’18\cite{bib64}} & {SynText800k,IC13,IC15,TT } &{71.8} &{4.8} & & {73.6} & {6.0}\\
{Two-stage\cite{bib65} }& {SynText800k,IC13,IC15,TT } &{45.0} & {-} & & {45.9} & {-}\\
{TextNet\cite{bib65} }& {SynText800k,IC13,IC15,TT}  & {54.0} & {2.7} & & {57.9} & {3.0}\\
{Li et al.\cite{bib66} }& {SynText840k,IC13,IC15,TT,MLT,AddF2k } & {57.8} & {1.4} & & {60.9} & {2.0}\\
{MaskTextSpotter’19\cite{bib67}} & {SynText800k,IC13,IC15,TT,AddF2k}  & {65.3} &{2.0} & & {66.9} & {2.4}\\
{Qin et al.\cite{bib68}} & {SynText200k,IC15,COCO-Text,TT,MLT}   & {67.8} & {4.8} & & {68.9} & {5.4}\\
{CharNet\cite{bib69}} &{ SynText800k,IC15,MLT,TT } &{ 66.2} & {1.2} & & {70.9} & {1.4}\\
{TextDragon\cite{bib70}} &{ SynText800k,IC15,TT}  & {48.8} &{ -}  & & {48.9} & {-}\\
{ABCNet\cite{bib13}} & {SynText150k,COCO-Text,TT,MLT}  & {74.2} & {22.8} & & 78.9 & 23.4\\
{PAN++\cite{bib73}} &{SynText150k,COCO-Text,TT,MLT}  & {77.1} &{ 24}  & & {81.5} & {25.1}\\
{FCENet\cite{bib74}} &{SynText150k,COCO-Text,TT,MLT}  & {77.5} &{ 24.1}  & & {82.1} & {25.3}\\
\hline
\end{tabular}
}
\end{table}
\subsubsection{Importance of Text Recognition Module}
To prove the effectiveness of the system, we evaluate our OCR module on both the public dataset and our Parking OCR dataset. As shown in Table \ref{table06}, the result of the public dataset shows FCENet achieves state-of-the-art performance, significantly outperforming all previous methods, especially in the running time. For the private Parking OCR dataset, the result (Table \ref{table3}) shows that the classification and recognition rate of these text instances have achieved 98.73$\%$ and 72.33$\%$ for the training and testing set respectively. While The anomaly filter can wipe out the rest of the 60.52$\%$  anomalies. The overall performance on the training set and testing set are 99.91$\%$ and 79.88$\%$ respectively. This means there are nearly 80$\%$ parking numbers' absolute locations that are accurate for the parking lots in the test set.
\begin{table}[]
\centering
\caption{Results of Text Recognition Module and Video-stream based Anomaly Filtering Module. "Detection", "Recognition" and "Classification" refer to the location, content, and class output of text instances respectively. "TP", "FP" and "FN" refer to True Positive, False Positive and False Negative respectively. Accuracy for Anomaly Filtering means the correctness rate of filtering out the anomaly. The Overall Recognition Accuracy indicates the recognition accuracy after the anomaly filter.}
\label{table3}
{
\begin{tabular}{l|ll}
\hline Text Recognition
  & Training set & Testing set     \\
  \hline
Detection TP  & 98.81\%  & 85.94\% \\
Detection FP  & 0.88\%   & 14.12\% \\
Detection FN  & 0.09\%   & 1.03\%  \\
Detection IOU & 95.94\%  & 86.08\% \\
Recognition TP  & 99.87\% & 82.49\% \\
Classification TP & 99.92\% & 84.16\% \\
\hline
\hline Anomaly Filtering
    & \multicolumn{2}{l}{\qquad  Unsupervised}     \\
\hline
Filter Accuracy &  \multicolumn{2}{l}{\qquad \quad 60.52\%} \\
\hline
Overall Recognition Accuracy & \textbf{99.91\%}  & \textbf{79.88\%} \\
\hline
\end{tabular}
}
\end{table}

\subsubsection{Importance of Anomaly Filtering Module}
\begin{figure}[htbp]
    \centering
    \includegraphics[width=12cm]{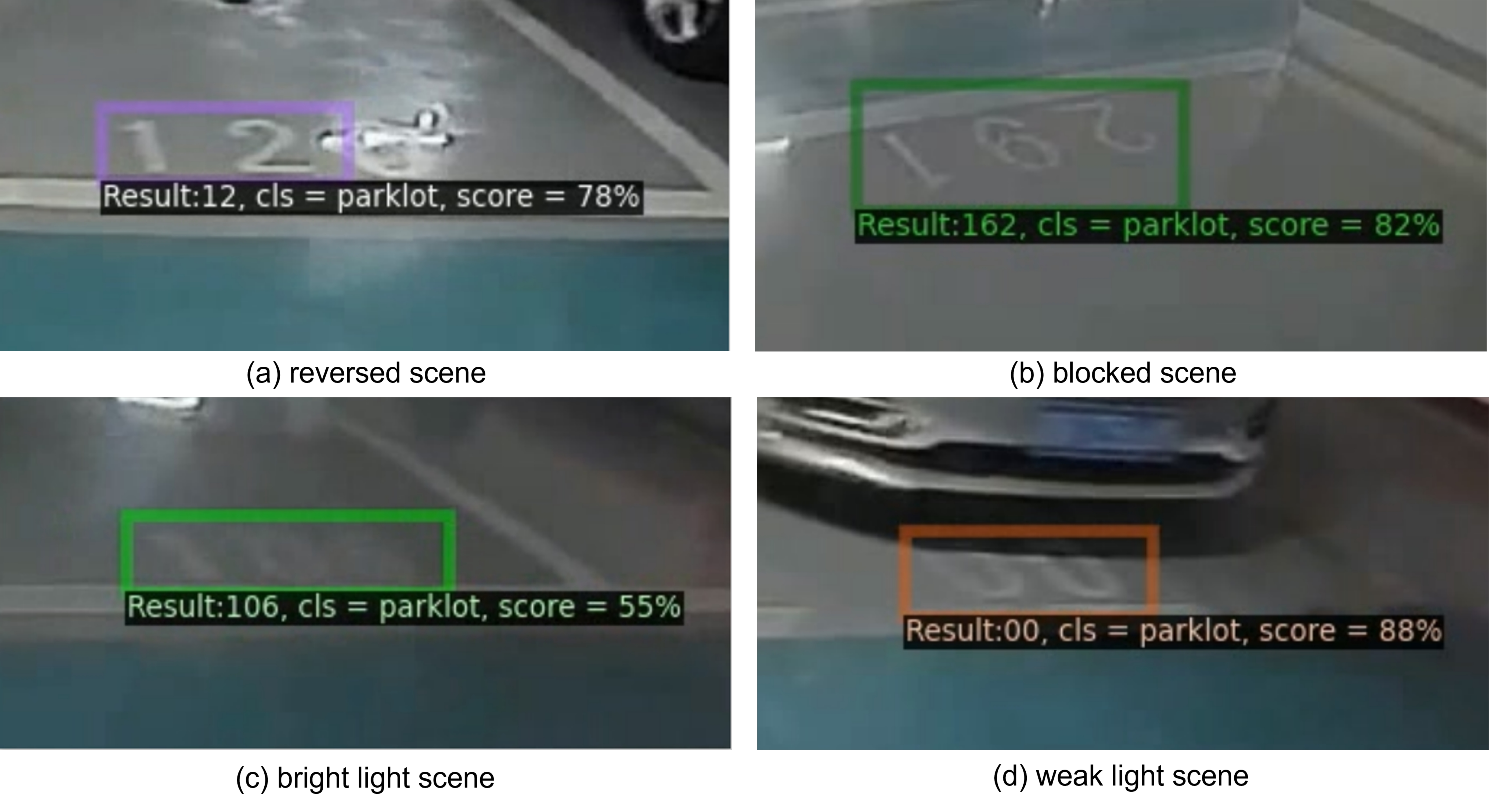}
    \caption{Four scenarios with identification.}
    \label{image8}
\end{figure}

Apart from the anomaly filtering module, the wrong text recognition results (Fig.\ref{image8}) will be further filtered out by their physical positions to improve the localization stability of the system. More specifically, the absolute 3D position of recognition text can be obtained by matching it with the HD semantic map. Subsequently, the test result will be removed if the distance between its physical position and the average position of texts in several previous frames exceeds the threshold. As shown in Fig.\ref{image7}, the localization errors are mostly less than 12cm along the x-axis and less than 15cm along the y-axis, which proves the localization accuracy of this system.

\subsubsection{Comparison with other positioning systems}
\begin{table}[htbp]
\begin{center}
\caption{Accuracy and latency comparison under both open (O) and isolated (I) areas. The GPS/RTK-based latency under isolated conditions is immeasurable since it is based on signal strength.}
\label{table05}       
\begin{tabular}{c|cccc}
\hline\noalign{\smallskip}
{Methods }& {Accuracy (I)} & {Latency (I) }&{ Accuracy (O) }& {Latency (O)} \\
\noalign{\smallskip}\hline\noalign{\smallskip}
{GPS} & $\geq${20m} & {immeasurable} & {10m} & {10ms} \\
{RTK }& $\geq${20m}  & {immeasurable} & $\leq${0.05m} & {10ms} \\
\hline\noalign{\smallskip}
{\textbf{Ours}} & $\leq${\textbf{0.3m}} & {\textbf{20ms}}  & {\textbf{0.3m}} & $\leq${\textbf{20ms}} \\
\hline
\end{tabular}
\end{center}
\end{table}
To fairly compare with other positioning systems, we follow the evaluation metrics proposed by \cite{bib24} that the metrics of a positioning system are composed as follows:

\begin{itemize}
\item Availability: The positioning technology can be used at the user's end and does not require special hardware.
\item Cost: The positioning system should reduce the cost of incurring additional infrastructure.
\item Energy Efficiency: The positioning system must not consume too much energy, such as the batteries of the user-side devices.
\item Reception Range: The positioning system can work in large spaces.
\item Accuracy: The ideal positioning system should be able to limit the positioning error to less than 10cm\cite{bib24}.
\item Latency: The positioning system should report the position without significant delay, which requires that the positioning system must not use too much reference information.
\item The positioning system should be able to locate multiple objects simultaneously or provide services to a large number of users in a large space at the same time.
\end{itemize}
\begin{table}[htbp]
  \begin{center}

  \small
  \caption{Comparison across different types of positioning systems, "W" stands for WIFI, "C" stands for Cellular, "B" stands for "Bluetooth", "US" stands for Ultrasound, "AS" stands for Acoustic Signals, 'RFID" stands for Radio Frequency Identification and "VL" stands for Visible Light.}
  \label{table4}
  \resizebox{\linewidth}{!}{
    \begin{tabular}{|p{10em}|p{2em}|p{5em}|p{2em}|p{5em}|p{4em}|p{4em}|p{4em}|p{4em}|}
    \toprule
    \textbf{System} & \textbf{Type} & \textbf{Availability} & \textbf{Cost} & \textbf{Energy Efficiency} & \textbf{Reception Range} & \textbf{Accuracy} & \textbf{Latency} & \textbf{Scalability} \\
    \midrule
    Hours & W  & $\surd$     & $\surd$     & ×     & $\surd$     & $\surd$     & $\surd$     & $\surd$ \\
    \midrule
    Redpin & W & $\surd$     & $\surd$     & $\surd$     & $\surd$     & ×     & ×     & $\surd$ \\
    \midrule
    BAT   & U & ×     & ×     & $\surd$     & $\surd$     & ×     & $\surd$     & $\surd$ \\
    \midrule
    Cricket & U & ×     & ×     & ×     & $\surd$     & $\surd$     & $\surd$     & $\surd$ \\
    \midrule
    Guoguo & AS & $\surd$     & ×     & ×     & ×     & $\surd$     & ×     & $\surd$ \\
    \midrule
    WalkieLokie & AS & $\surd$     & ×     & ×     & ×     & $\surd$     & ×     & $\surd$ \\
    \midrule
    iBeacon based system in  & B & $\surd$     & ×     & ×     & $\surd$     & ×     & ×     & $\surd$ \\
    \midrule
    RF-Compass & RFID  & ×     & $\surd$     & $\surd$     & $\surd$     & $\surd$     & $\surd$     & $\surd$ \\
    \midrule
    PinIt & RFID  & ×     & $\surd$     & $\surd$     & $\surd$     & ×     & ×     & $\surd$ \\
    \midrule
    LocaLight & VL & $\surd$     & $\surd$     & $\surd$     & ×     & $\surd$     & ×     & × \\
    \midrule
    \textbf{Ours} & \textbf{VL} & \textbf{×} & \textbf{×} & \textbf{$\surd$} & \textbf{$\surd$} & \textbf{$\surd$} & \textbf{$\surd$} & \textbf{$\surd$} \\
    \bottomrule
    \end{tabular}
    }
  \end{center}
\end{table}%

Regarding the metrics, the result is shown in Table \ref{table4}. Our system outperforms others with a good trade-off between the performance phase(reception range, accuracy, latency, and scalability.) and the cost phase(availability, cost, and energy efficiency). Regarding its performance, since it does not require any additional device remotely, it works anywhere(wide reception range) with high accuracy and low latency and can easily deploy to other vehicles(scalability) as long as the system is under the parking scene. Although our system requires the autonomous hardware to be installed, which will fail the cost and availability(ready for any users), the autonomous system will be widely deployed in the future and reduce the financial\&using-cost subsequently.
\begin{figure}[htbp]
\centering
\includegraphics[height=4cm,width=5cm]{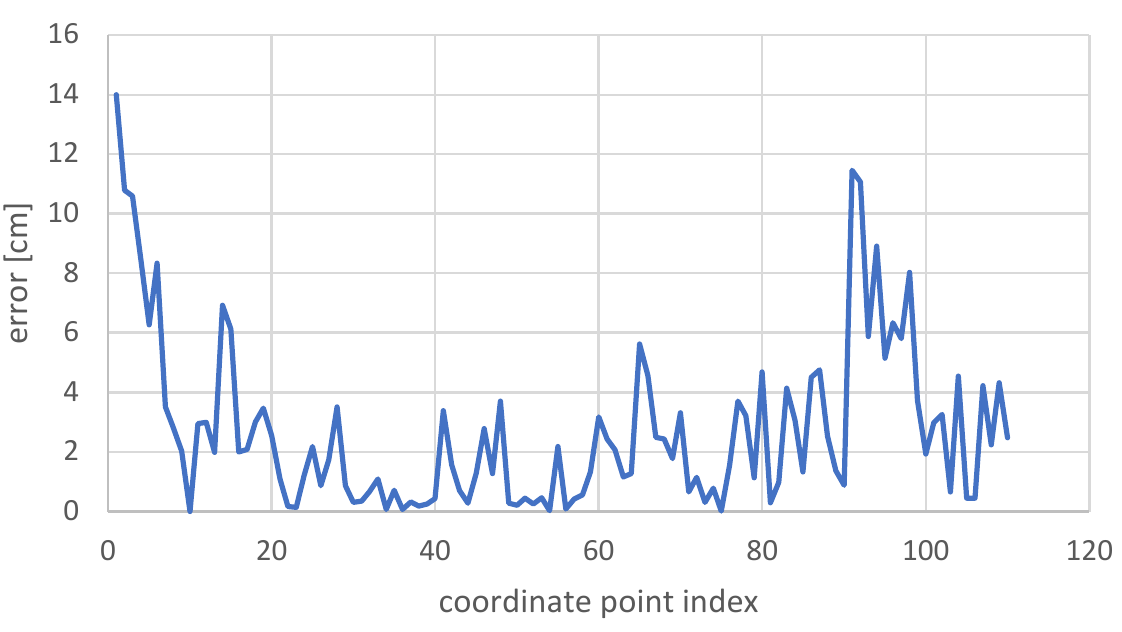}
\includegraphics[height=4cm,width=5cm]{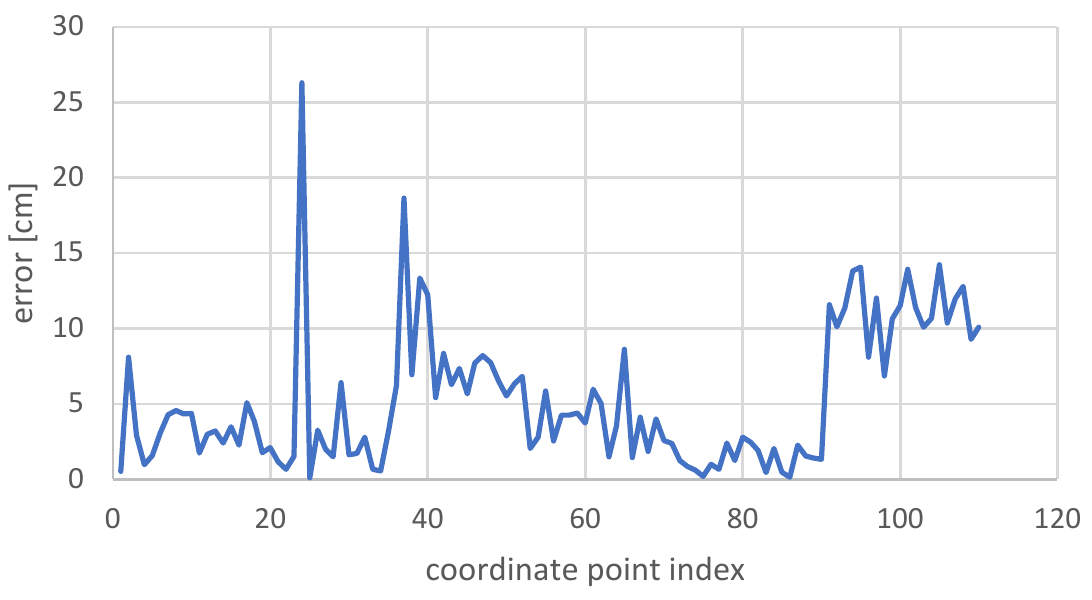}
\caption{Localization error on our method. (left) The localization error along the x-axis direction. (right) The localization error along the y-axis direction.}
\label{image7}
\end{figure}
We also compare the positioning accuracy under both isolated and open scenes using GPS and RTK with our method in Table \ref{table05}. For the isolated scene, due to the weak signal from the underground parking lot, the precision of both GPS and RTK is greater than 20 meters while Fig. \ref{image7} shows the accuracy of our system that the average offset of both horizontal(x) and vertical(y) axis of ego vehicle are within 10cm, and the overall positioning offset is less than 0.3 meters. For the open area, The GPS's precision keeps a great positioning offset of 10 meters while RTK and our method have a precision of 0.05 meters and 0.3 meters respectively. 

\subsubsection{Comparison experiments with different isolated conditions}

To further prove the superiority of our positioning system in the isolated environment, we list some of the performance of some complicated conditions and compare our positioning system under these conditions. It can be seen from Table \ref{table8} that our positioning system can also show superior positioning capabilities under complex conditions, with less than 5cm positioning errors.

\begin{table}[htb]
\caption{Comparison experiments with different isolated conditions based on our system. The positioning error of bright light, weak light, blur font, 45-degree rotation font, 90-degree rotating font, and 180-degree rotating font (accurate to cm).}
\label{table8}
\resizebox{\linewidth}{!}{ 
\begin{tabular}{cccccc}
\hline
 Bright light & Weak light & Fuzzy  & 45-degrees rotation  & 90-degrees rotation  & 180-degrees rotation  \\ \hline
 5cm         & 5cm       & 4cm       & 3cm                & 3cm                & 3cm                 \\ \hline
\end{tabular}}
\end{table}

\section{Discussion}
In this article, we proposed an OCR-based positioning system aimed at the isolated environment. it can help the ego car to localize itself and gives a much more accurate position than others positioning systems while the wireless signal is invalid. However, the shortages of this system are also significant in that it only works when The following conditions are satisfied:
\begin{itemize}
\item The map of the parks is established.
\item there are parking numbers available in the parking lots.
\end{itemize}
Both constraints limit the usage of this system. Regarding the constraints, future explorations can be making OCR satisfy other scenarios such as high speeding way, this system could detect and recognize the signs on the road and localize itself roughly by combing the recognition and map information.

\section{Conclusions}

In this paper, we propose an integrated real-time vehicle localization system by detecting and recognizing the parking number. Specifically, we employ the text recognition module to localization and classification the parking number. In addition, we utilize a video-stream-based anomaly filtering module to filter incorrect recognition. Furthermore, we propose a localization projection module to transform the detection result to a related location. Extensive results show that the proposed solution can improve the performance on accuracy and stability simultaneously. In particular, our solution achieves state-of-the-art performance on the Parking OCR Dataset.

\section*{Declarations}

No potential conflict of interest was reported by the authors.

\end{document}